\documentclass{ieeeaccess}
\usepackage{cite}
\usepackage{amsmath,amssymb,amsfonts}
\usepackage{algorithmic}
\usepackage{graphicx}
\usepackage{comment}
\usepackage{textcomp}
\def\BibTeX{{\rm B\kern-.05em{\sc i\kern-.025em b}\kern-.08em
    T\kern-.1667em\lower.7ex\hbox{E}\kern-.125emX}}

\usepackage{xcolor}
\usepackage{pict2e}
\usepackage{color}
\usepackage{multirow}

\usepackage{enumitem}
\newlist{steps}{enumerate}{1}
\setlist[steps, 1]{label = \textbf {Step \arabic*}:, leftmargin=1.40cm}

\newsavebox{\ORCIDlogo}
\savebox{\ORCIDlogo}{%
\setlength{\unitlength}{\dimexpr 1em/256\relax}%
\begin{picture}(256,256)%
  \color[HTML]{A6CE39}\put(128,128){\circle*{256}}%
  \color{white}%
  \put(78.6,199.2){\circle*{20}}%
  \moveto(70.9,176,9)\lineto(86.3,176,9)\lineto(86.3,69.8)\lineto(70.9,69.8)%
  \closepath\fillpath%
  \moveto(108.9,176.9)\lineto(150.5,176.9)%
  \curveto(190.1,176.9)(207.5,148.6)(207.5 ,123.3)%
  \curveto(207.5,95,8)(186,69.7)(150.7,69.7)%
  \lineto(108.9,69.7)%
  \closepath\fillpath%
  \color[HTML]{A6CE39}%
  \moveto(124.3,83.6)\lineto(148.8,83.6)%
  \curveto(183.7,83.6)(191.7,110.1)(191.7,123.3)%
  \curveto(191.7,144.8)(178,163)(148,163)%
  \lineto(124.3,163)%
  \closepath\fillpath%
\end{picture}%
}

\newcommand\orcidicon[1]{\href{https://orcid.org/#1}{\usebox{\ORCIDlogo}}}

\usepackage{hyperref} 
\hypersetup{
  colorlinks=false,
  linkbordercolor=white,
 urlbordercolor=white,
pdfborder={0 0 0}
} 
\usepackage[switch,columnwise]{lineno}
\begin{document}
\history{Date of publication xxxx 00, 0000, date of current version xxxx 00, 0000.}
\doi{10.1109/ACCESS.2022.0092316}
\title{Machine Learning-Based Differential Diagnosis of Parkinson's Disease Using Kinematic Feature Extraction and Selection}

\author{ \uppercase{Masahiro Matsumoto\authorrefmark{1}},  \uppercase{Abu Saleh Musa Miah \orcidicon{0000-0002-1238-0464} \authorrefmark{1},
\IEEEmembership{Member, IEEE}}, \uppercase{Nobuyoshi Asai \authorrefmark{1}},
 and \uppercase{Jungpil Shin \orcidicon{0000-0002-7476-2468} \authorrefmark{1},
\IEEEmembership{Senior Member, IEEE}}}

\address[1]{School of Computer Science and Engineering, The University of Aizu, Aizuwakamatsu 965-8580, Japan, abusalehcse.ru@gmail.com, ma2mo10ma3hi@gmail.com}

\tfootnote{This work was supported by the Competitive Research Fund of The University of Aizu, Japan.}
\markboth
{....}
{This paper is currently under review for possible publication in IEEE Access.}

\corresp{Corresponding author: Jungpil Shin (e-mail: jpshin@u-aizu.ac.jp)}


\begin{abstract}Parkinson's disease (PD), the second most common neurodegenerative disorder, is characterized by dopaminergic neuron loss and the accumulation of abnormal synuclein. PD presents both motor and non-motor symptoms that progressively impair daily functioning. The severity of these symptoms is typically assessed using the MDS-UPDRS rating scale, which is subjective and dependent on the physician's experience. Additionally, PD shares symptoms with other neurodegenerative diseases, such as progressive supranuclear palsy (PSP) and multiple system atrophy (MSA), complicating accurate diagnosis.
To address these diagnostic challenges, we propose a machine learning-based system for differential diagnosis of PD, PSP, MSA, and healthy controls (HC). This system utilizes a kinematic feature-based hierarchical feature extraction and selection approach. Initially, 18 kinematic features are extracted, including two newly proposed features: Thumb-to-index vector velocity and acceleration, which provide insights into motor control patterns. In addition, 41 statistical features were extracted here from each kinematic feature, including some new approaches such as Average Absolute Change, Rhythm, Amplitude, Frequency, Standard Deviation of Frequency, and Slope. 
Feature selection is performed using One-way ANOVA to rank features, followed by Sequential Forward Floating Selection (SFFS) to identify the most relevant ones, aiming to reduce the computational complexity. The final feature set is used for classification, achieving a classification accuracy of 66.67\% for each dataset and 88.89\% for each patient, with particularly high performance for the MSA and HC groups using the SVM algorithm. This system shows potential as a rapid and accurate diagnostic tool in clinical practice, though further data collection and refinement are needed to enhance its reliability.
\end{abstract}

\titlepgskip=-15pt
\maketitle
\section{Introduction}
Parkinson’s disease (PD) is the second most common neurodegenerative disease in the world after Alzheimer's disease. It is caused by the degeneration of midbrain dopaminergic neurons and the accumulation of Lewy bodies composed of abnormal synuclein within the brain. PD has both motor and non-motor symptoms, which gradually worsen and significantly impact daily life. In particular, motor symptoms are characterized by a clinical triad of muscle rigidity, tremor, and postural reflex disturbances, which are key to diagnosing the disease \cite{poewe2017parkinson,jankovic2008parkinson,postuma2015mds}. 
There are many clinical assessment methods to evaluate PD symptoms beside other disease \cite{shin2024parkinson_hrooka_miah,10707606_miah_eye_disease,hassan2024residual_miah_ad_najm,siddiqua2024computer_miah_knee_disease}, including Hoehn-Yahr staging, the Schwab and England rating, and the Unified Parkinson's Disease Rating Scale (UPDRS). In particular, the MDS-UPDRS, created by the Movement Disorder Society (MDS) in 2008 to revise the UPDRS, is widely used. The MDS-UPDRS is divided into four sections, with part III assessing motor symptoms \cite{goetz2008movement}. In this section, the patient performs a motor task and is rated on a 5-point scale of 0 to 4 by the diagnosing physician visually. For example, finger tapping (FT) tasks are evaluated visually for speed, magnitude, and rhythm to assess akinesia. Therefore, this diagnosis requires a physician who is well-versed in the field, and there is variation in the evaluation among physicians.
In addition, other neurodegenerative diseases can also present with symptoms similar to PD, making them difficult to differentiate, especially in the early stages \cite{shi2013gray}. These diseases, known as atypical Parkinsonism (AP), include progressive supranuclear palsy (PSP) and multiple system atrophy (MSA). Despite their symptom similarity to PD, they have been reported to be less responsive to the antiparkinsonian drug levodopa \cite{lang2005treatment,ishida2021effectiveness}. The diagnostic criteria for these diseases include differences in treatment responsiveness, major symptoms, age of onset, and symptom progression \cite{constantinescu2007levodopa}. Characteristic symptoms of PSP include postural instability, eye movement disorders, and cognitive dysfunction. Characteristic symptoms of MSA include cerebellar ataxia, slow movement, and autonomic dysfunction. However, a 2015 study observed 134 patients clinically diagnosed with MSA and found that the correct diagnosis was made at autopsy in only 83 patients (62\%); MSA was most often misdiagnosed as Lewy body dementia, followed by PSP and PD \cite{koga2015dlb}. Therefore, it is crucial to apply accurate and rapid measures of motor symptoms in PD and AP in order to provide effective treatments for each disease.
Recently, many methods for PD detection and differential diagnosis have been developed by integrating devices and AI \cite{dennis2024role}. For example, MRI and CT images are being used for diagnosis by AI \cite{salvatore2014machine,chougar2024contribution,archer2019development,chien2020using}. However, due to cost and time requirements, they may not be suitable for practical application in clinical settings. In addition, studies addressing the problem of differential diagnosis between PD and AP are generally few, as most studies are based on imaging\cite{song2022differential,daoudi2022speech,belic2023quick}.
The objective of this study was to create a system that is lightweight, low-cost, and can be used in a clinical setting for rapid diagnosis. The proposed system performs differential diagnosis by using a modified version of the FT task in MDS-UPDRS III. While this test has traditionally relied on the subjective judgment of the diagnosing physician, the proposed system used an inertial sensor to record FT task and analyzed the data obtained to automate the differential diagnosis of hand motor symptoms. Our major contribution is given below:
\begin{itemize}
    \item \textbf{New Kinematic Feature Extraction Method:} 
    we propose a machine learning-based system for differential diagnosis of PD, PSP, MSA, and healthy controls (HC). This system utilizes a kinematic feature-based hierarchical feature extraction and selection approach. Initially, 18 kinematic features are extracted, including two newly proposed features: Thumb-to-index vector velocity and acceleration, which provide insights into motor control patterns. 
    These features are derived from the motion of the thumb and index finger during a finger tapping (FT) task. These features were calculated based on 6 attributes (X, Y, Z) for both the thumb and index finger, along with 12 additional features obtained through numerical differentiation and vector magnitude. The 18 kinematic features were divided into three categories: Angular Velocity (N = 6), Angular Acceleration (N = 6), and Vector Magnitude (N = 6). This set of features captures detailed motion characteristics such as rotational speed, acceleration, and the relative motion between the thumb and index finger, providing a deeper understanding of motor control. 

    \item \textbf{Hierarchical Feature Extension:} 
    To enhance the kinematic features, then 41 statistical features were extracted here from each kinematic feature, including some new approaches such as Average Absolute Change, Rhythm, Amplitude, Frequency, Standard Deviation of Frequency, and Slope. 
    These statistical methods are used to generate hierarchical features that provide additional insights into the writing process. These features, also included central tendency, dispersion, and higher-order relationships, included new metrics. The integration of these statistical features with the kinematic features resulted in a comprehensive feature set of 758 features (18 kinematic features × 41 statistical features). This extended feature set proved highly effective in identifying PD-specific movement patterns, enhancing the system's ability to differentiate PD from other movement disorders like Progressive Supranuclear Palsy (PSP) and Multiple System Atrophy (MSA).

    \item \textbf{Feature Selection with Sequential Forward Floating Selection (SFFS):} 
    We employed One-way ANOVA to select significant features, setting a threshold of \textit{p}-values smaller than 0.005. Using this method, we refined the feature set by selecting the most relevant features that contributed to optimal classification performance. The Sequential Forward Floating Selection (SFFS) algorithm was then applied to identify the most impactful features, ensuring the robustness of the feature set across various machine learning classifiers. This approach significantly improved the detection of PD, PSP, MSA, and healthy controls (HC), as it enabled the model to focus on the most discriminative features.

    \item \textbf{Classifier Optimization:} 
    The optimized feature set was used to train multiple machine learning classifiers, with the Support Vector Machine (SVM) yielding the highest performance. We achieved a classification accuracy of 66.67\% for each data point and 88.89\% for each patient, with particularly strong results for the MSA and HC groups. However, some PD and PSP patients were occasionally misclassified. This highlights the strength of our approach to PD detection and sets a new benchmark in movement disorder classification methodologies.
\end{itemize}

The remainder of this paper is organized as follows: Section \ref{sec:Literature Review} reports on related work. Section \ref{sec:Dataset} describes the baseline characteristics of 54 patients and the class distribution in the public data. Section \ref{sec: Materials and methods} describes the proposed methodology, feature extraction, and feature selection using One-way ANOVA and SFFS. Section \ref{sec: Experimental Setup and Performance Metrics} presents the experimental setup, performance metrics, and the experimental results. Finally, the conclusion and future work are presented in Section \ref{sec: Conclusion and future plan}.

\section{Related Work}
\label {sec:Literature Review} 
In recent years, many studies have integrated devices and algorithms to evaluate the severity of motor symptoms and differentiate diseases based on kinematic characteristics of various movements. Belic et al. developed a wearable system that uses inertial sensors \cite{10704663_miah_har_sensor} to perform differential diagnosis from repeated finger tapping movements \cite{belic2023quick}. In this system, five signal transformations were performed from the acceleration data obtained by the sensor: 1) raw signal representing angular velocity, 2) integral representing finger swing angle, 3) derivative representing angular acceleration, 4) square of the signal, and 5) Fourier transform representing frequency content of the signal. In this study, FT movements were recorded using sensors from 14 PD patients, 16  PSP - Richardson's syndrome (PSP-RS) patients, 13 MSA of predominantly parkinsonian type patients, and 11 HC patients, yielding a classification accuracy of 85.18\%.
Song et al. developed a classification system based on postural instability and gait analysis that can differentiate PD, ataxia, MSA - cerebellar subtype (MSA-C), and PSP-RS\cite{song2022differential}. They employed the Enhanced Weight Voting Ensemble (EWVE) method to learn the characteristics of each disease by combining two classifiers for gait and postural instability. This allowed them to avoid errors caused by a single classifier and improve classification accuracy. The AI model used in the gait analysis was based on a machine learning method called gated recursive units (GRUs), while the AI model used in the postural instability was based on deep neural networks (DNN). In this study, 551 cases of PD, 38 cases of PSP-RS, 113 cases of cerebellar ataxia, and 71 cases of MSA-C were used for differential diagnosis between PD and each AP. The results showed that the AUC for PD vs. ataxia patients was 0.974$\pm$0.036, sensitivity 0.829$\pm$0.217, specificity 0.969$\pm$0.038, AUC for PD vs. MSA-C was 0.975$\pm$0.020, sensitivity 0.823$\pm$0.162, specificity 0.932$\pm$0.030, AUC for PD vs. PSP-RS 0.963$\pm$0.028, sensitivity 0.555$\pm$0.157, and specificity 0.936$\pm$0.031.
Salvatore et al. created a model to differentiate midbrain, thalamus, cortex, and corpus callosum differences between PD (n = 28) and PSP (n = 28) and healthy subjects (n = 28) from T1-weighted MRI\cite{salvatore2014machine}. Principal Components Analysis (PCA) was used for feature extraction, and SVM was used as the classification algorithm. The model to differentiate PSP from PD showed 88.9\% accuracy. In addition,  The model to differentiate PSP subjects from HC showed an accuracy of 89.1\%.
The aim of our study was to create a highly accurate and simplified diagnostic system using a lightweight, low-cost device for use in clinical practice. A differential diagnosis was performed using the data from Belic et al.'s SVM study.

\section{Dataset}
\label {sec:Dataset} 
\begin{figure}[]
\centering
\includegraphics[width=2.5  in]{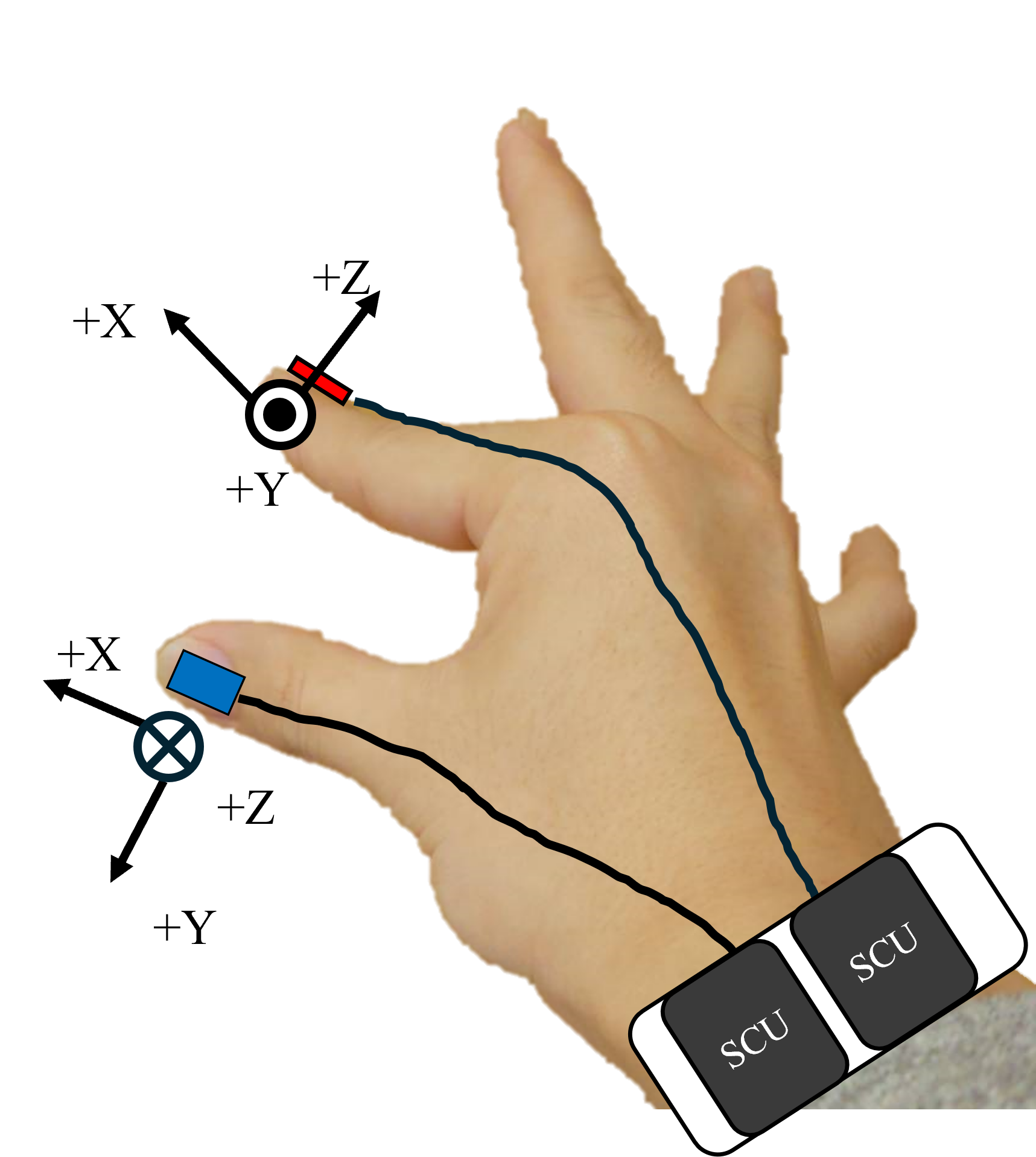}
\caption{Sensor Control Unit coordinate system.}
\label{Figure:1}
\end{figure}

We used publicly available data on the FT task, which were collected in Belic's study \cite{belic2023quick} and deposited on GitHub. 267 samples recorded via a gyroscope were made available. These data samples capture the Finger Tapping task performed by 54 patients using their more symptomatic right hand. Each patient conducted the task up to six times, depending on their specific condition and tolerance. The data collection device used in Belic's study was a wearable gyroscope, SCU, which is shown in Figure \ref{Figure:1}. SCU can acquire signal data in the thumb and index finger and then transmit it wirelessly to a computer, and the data acquisition software has an easy-to-use GUI for any user. In addition, wireless communication covers a 20-meter radius indoors, allowing for easy use of the system in a clinical environment. Furthermore, IMU is a light weight and compact size, so it allows subjects to perform movements naturally. 

\begin{figure*}[]
\centering
\includegraphics[width=7in]{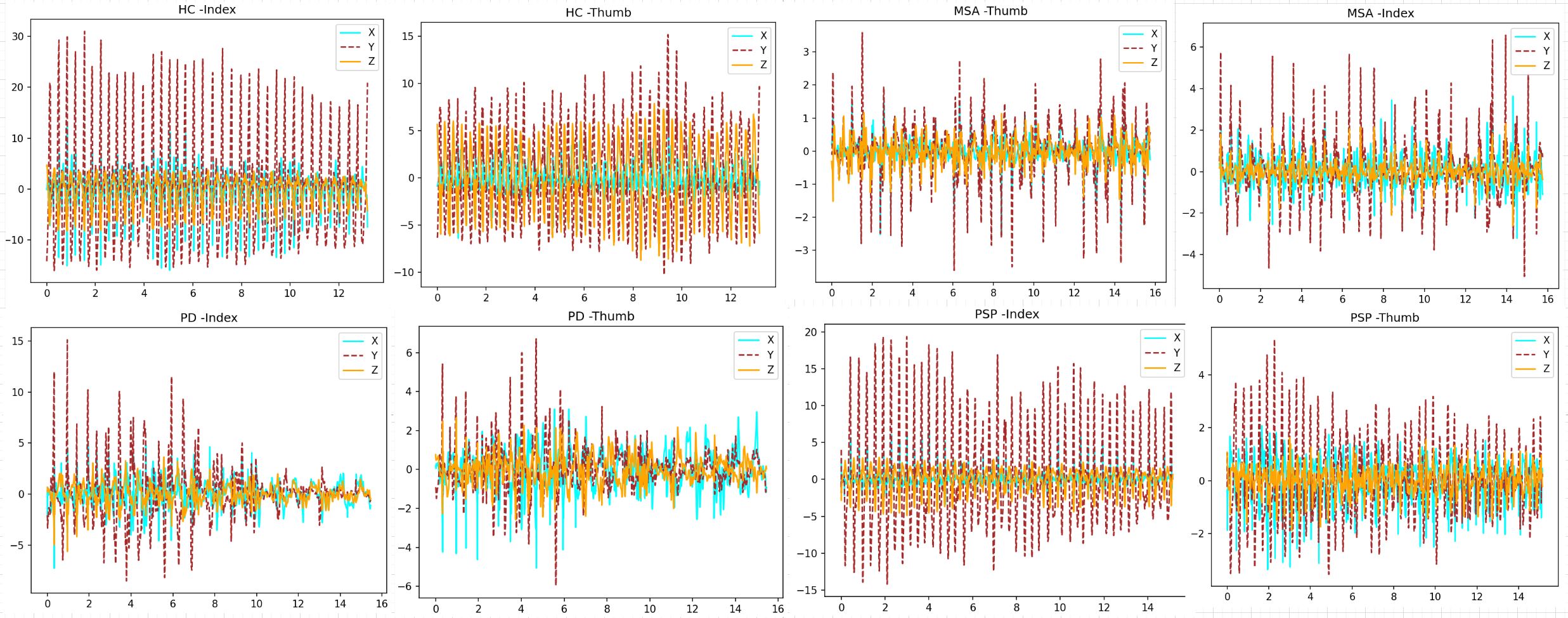}
\caption{Example of the Dataset Signal for Individual Disease.}
\label{Figure:2}
\end{figure*}

The data are given in .mat file (matlab format) and contain the following information: Symptom ('CTRL', 'PD', 'MSA', or 'PSP'), '$gyroThumb\_\{X,Y,Z\}$' : x, y, z axes of the gyroscope in the thumb, '$gyroIndex\_\{X,Y,Z\}$': x, y, z axes of the gyroscope in the index finger, 'personID': person code, 'trialID': trial code, Sampling rate (Hz) (but always 200 Hz). In this study, this .mat file was converted to a csv file and the triaxial gyroscope data $(gyroThumb\_\{X,Y,Z\}$, $gyroIndex\_\{X,Y,Z\})$ was used in the analysis. Figure \ref{Figure:2} shows examples of dataset signals for each disease.

\subsection{Baseline characteristics of Each Patient}
Data were collected from 14 PD patients (mean age 62.1$\pm$9.4 years, 71.4\% male), 16 PSP patients (mean age 67.1$\pm$8.9 years, 68.8\% male), 13 MSA patients (mean age 58.4$\pm$4.8 years, 30.8\% male), and 11 healthy controls (mean age 55.9$\pm$8.4 years, 27.3\% male). Detailed clinical and demographic data are shown in Table \ref{tab:Table 6}. There were no significant differences in age or gender of the overall patient group.

\subsection{Class Distribution of Patients}
We analyzed triaxial gyroscopic data from PD, PSP, and MSA patients and HC and classified them into four categories. Each patient performed the FT task with the right hand, at most 6 trials. Distributions in these categories are shown in Table \ref {tab:Table 7}.

\begin{table*}
\centering
\caption{Baseline characteristics of each patient \label {tab:Table 6} } 
\begin{tabular}{llllllll}
\hline
    & Age (mean$\pm$SD) & Gender (F/M) & Disease duration (mean$\pm$SD) & Hoehn \& Yahr stage & UPDRS total & UPDRS III \\ \hline
PD  & 62.1 $\pm$ 9.4 & 4/10 & 4.9 $\pm$ 4.5 & 2.2 $\pm$ 0.8 & 48.1 $\pm$ 18.7 & 27.0 $\pm$ 9.8 \\ 
PSP & 67.1 $\pm$ 8.9 & 5/11 & 5.2 $\pm$ 2.3 & 3.8 $\pm$ 0.8 & 79.9 $\pm$ 17.2 & 45.7 $\pm$ 10.4 \\ 
MSA & 58.4 $\pm$ 4.8 & 9/4  & 3.47 $\pm$ 1.5 & 3.2 $\pm$ 0.7 & 77.2 $\pm$ 12.7 & 45.4 $\pm$ 8.6 \\ 
HC  & 55.0 $\pm$ 8.4 & 8/3  & - & - & - & - \\ 
HC-MSA & - & - & - & - & - & - \\ 
HC-PD & - & - & - & - & - & - \\ 
HC-PSP & p = 0.02 & - & - & - & - & - \\ 
MSA-PD & - & p = 0.05 & - & p < 0.01 & - & - \\ 
MSA-PSP & - & p = 0.02 & - & - & - & - \\ 
PD-PSP & - & - & - & p < 0.001 & - & - \\ \hline
\end{tabular} \\
\footnotesize Note: \textit{p}-values were obtained from One-way ANOVA tests and and Kruskal-Wallis one-way analysis for the age, sex, and UPDRS score variables, respectively.
\end{table*}

\begin{table}
\centering
\caption{Class distribution of patients. \label {tab:Table 7} }
\begin{tabular} {llllllll}

\hline
Disease & PD  & PSP  & MSA & HC  \\
\hline

Number of People               & 14  & 16  & 13 & 11 \\
Number of Data               & 67  & 76  & 72  & 52 \\

\hline
\end{tabular}
\end{table}



\section{Propsoed Method}
\label {sec: Materials and methods}
\begin{figure*}[ht]
\centering
\includegraphics[width=7.4in]{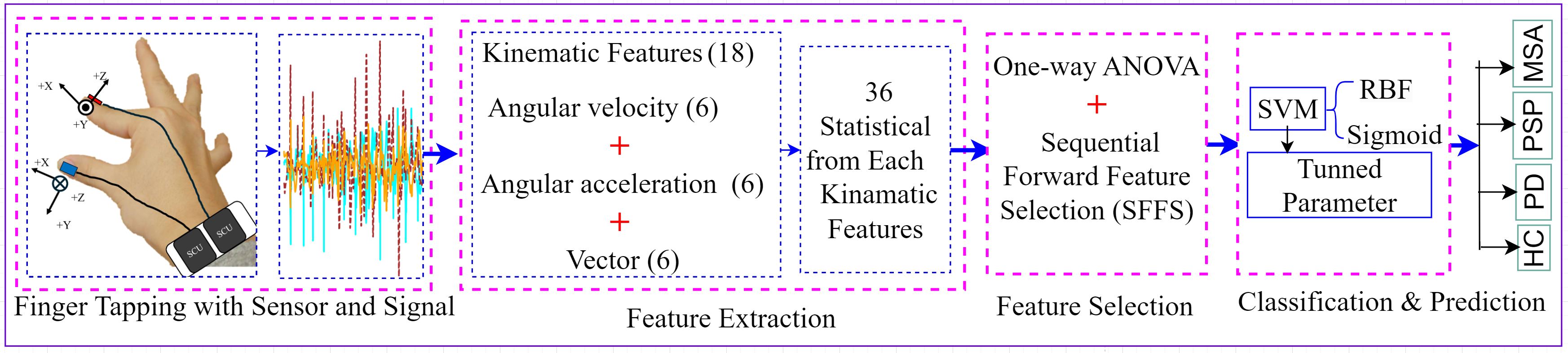}
\caption{Flowchart of the proposed ML-based method for differential diagnosis.}
\label{Figure:3}
\end{figure*}

The overall experimental workflow employed in this study is presented in Fig. \ref{Figure:3}. We conducted the experiment through a series of four key steps: feature extraction, feature selection, model training, and performance evaluation.
\begin{itemize}
\item 
\textbf{Step 1: Hierarchical Feature Extraction}
Hierarchical feature extraction captures the complex relationships among the features. In this study, we calculated hierarchical features by extracting statistical features from the kinematic features. There were 18  kinematic features were extracted, including two newly proposed features: Thumb-to-index vector velocity and acceleration, which provide insights into motor control patterns. The 18 kinematic features are shortly given below:  
\begin{itemize}
    \item \textbf{Raw Signal (Angular Velocity):} The angular velocity was directly computed from the movement signals of the thumb and index finger along the X, Y, and Z axes.
    \item \textbf{Angular Acceleration (Differentiation):} Angular acceleration was derived by numerically differentiating the angular velocity signals.
    \item \textbf{Vector Magnitude:} The magnitude of the relative movement vector between the thumb and index finger was calculated using the components of angular velocity and acceleration.
    \end{itemize}

In addition, 41 statistical features were extracted here from each kinematic feature, including some new approaches such as Average Absolute Change, Rhythm, Amplitude, Frequency, Standard Deviation of Frequency, and Slope. The statistical features were categorized into several groups based on the type of information they captured. Descriptive statistics include measures like $RMS$, $Min$, $Max$, $Avg$, $Std$, $Med$, and $Amplitude$, which describe the overall signal magnitude and variability. Spectral features such as $Max\_freq$, $Centroid$, $Frequency$, and $Frequency\_Std$ characterize the frequency domain of the signal. Peak and maximum features ($RMS\_Max$, $Min\_Max$, $Max\_Max$, $Avg\_Max$, $Std\_Max$, $Med\_Max$) focus on the extrema and statistical properties of peak points. New features included here Noise and signal quality features, including $Noise\_Var$, $SNR$, and $Conv\_Ene$, assess the signal’s strength and noise levels. Time-series and autocorrelation features like $Variance$, $Avg\_Abs\_Change$, $Autocorr\_Lagi$, and $Quant\_i$ capture temporal trends and correlations in the signal. $Rhythm$ quantifies the regularity of signal oscillations, while the $Slope$ feature measures the rate of change in the signal. These statistical features provided a detailed characterization of the movement patterns for each subject.
\item 
\textbf{Step 2: Feature Selection}
In the second step, we applied feature selection to identify the most relevant features for classification. We first calculated the \textit{p}-values for all the extracted features using One-way ANOVA. Features with \textit{p}-values smaller than 0.005 were considered statistically significant. Next, we employed Sequential Forward Floating Selection (SFFS) to refine the feature set and select the most impactful features based on their contribution to classification accuracy. This step ensured that only the most relevant features were retained for further analysis, thus reducing dimensionality and improving the model's efficiency.
 \item 
\textbf{Step 3: Model Training and Evaluation}
In the third step, we trained a Support Vector Machine (SVM) model using the selected feature set. Hyperparameter optimization was performed to fine-tune the SVM parameters. The Optuna framework was used to optimize three key SVM parameters: kernel function, cost, and gamma ($\gamma$). The optimization process was repeated 200 times, with a fixed seed value of 42 to ensure reproducibility. The optimal parameter values were selected based on the highest classification accuracy achieved during optimization. Finally, we evaluated the performance of the optimized SVM model using leave-one-subject-out cross-validation (LOOCV). This cross-validation technique ensured that the model was evaluated on each subject's data while training on data from the remaining subjects. The model was then used to classify each subject into one of the four categories: Healthy Control (HC), Parkinson's Disease (PD), Progressive Supranuclear Palsy (PSP), and Multiple System Atrophy (MSA). The performance of the model was assessed using classification accuracy, recall, precision, and F1-score.
\end{itemize}

\begin{figure*}[ht]
\centering
\includegraphics[width=4.7in]{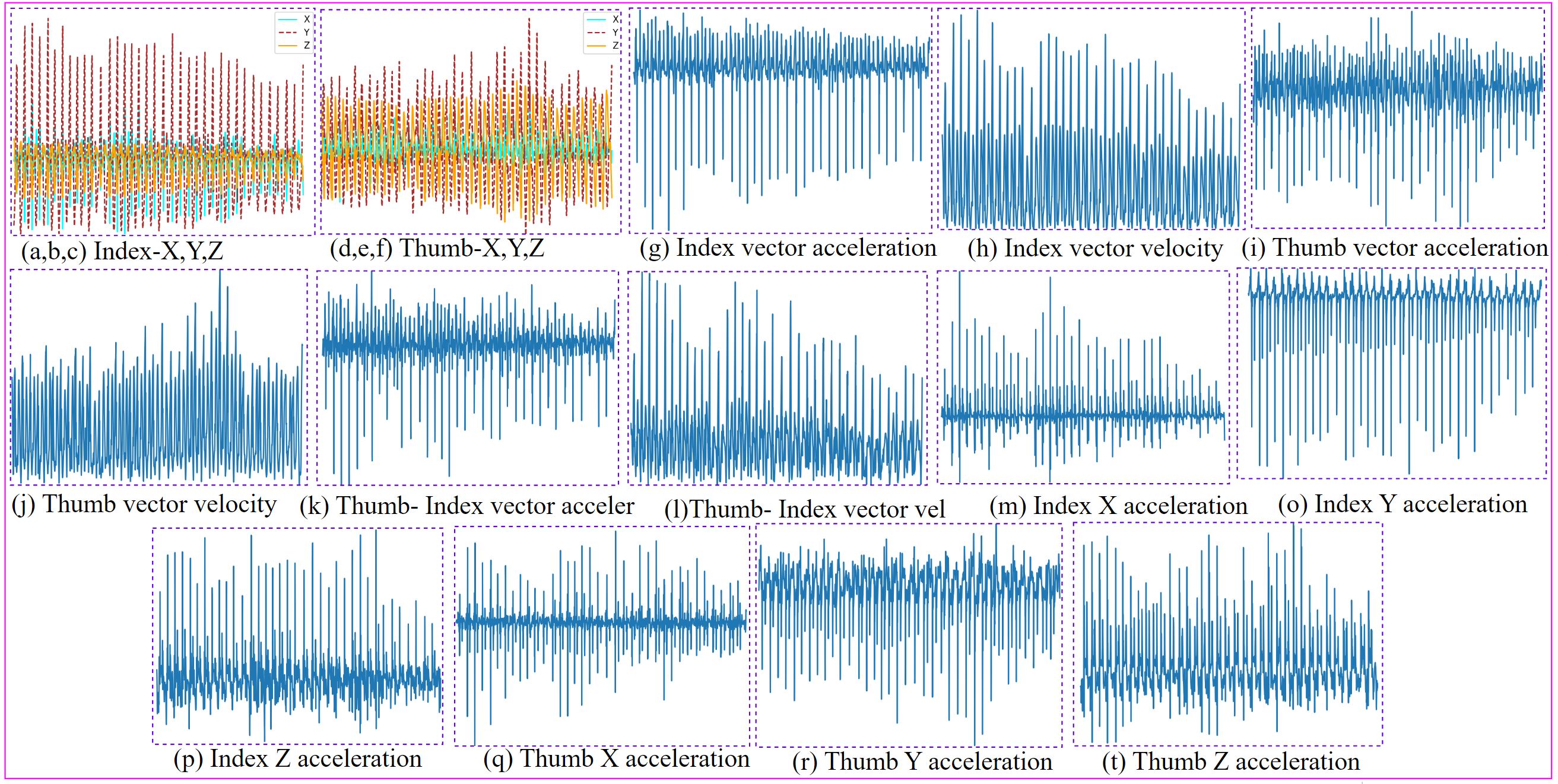}
\caption{Kinematic 18 feature for HC labels.}
\label{Figure:4}
\end{figure*}

\begin{figure*}[ht]
\centering
\includegraphics[width=4.7in]{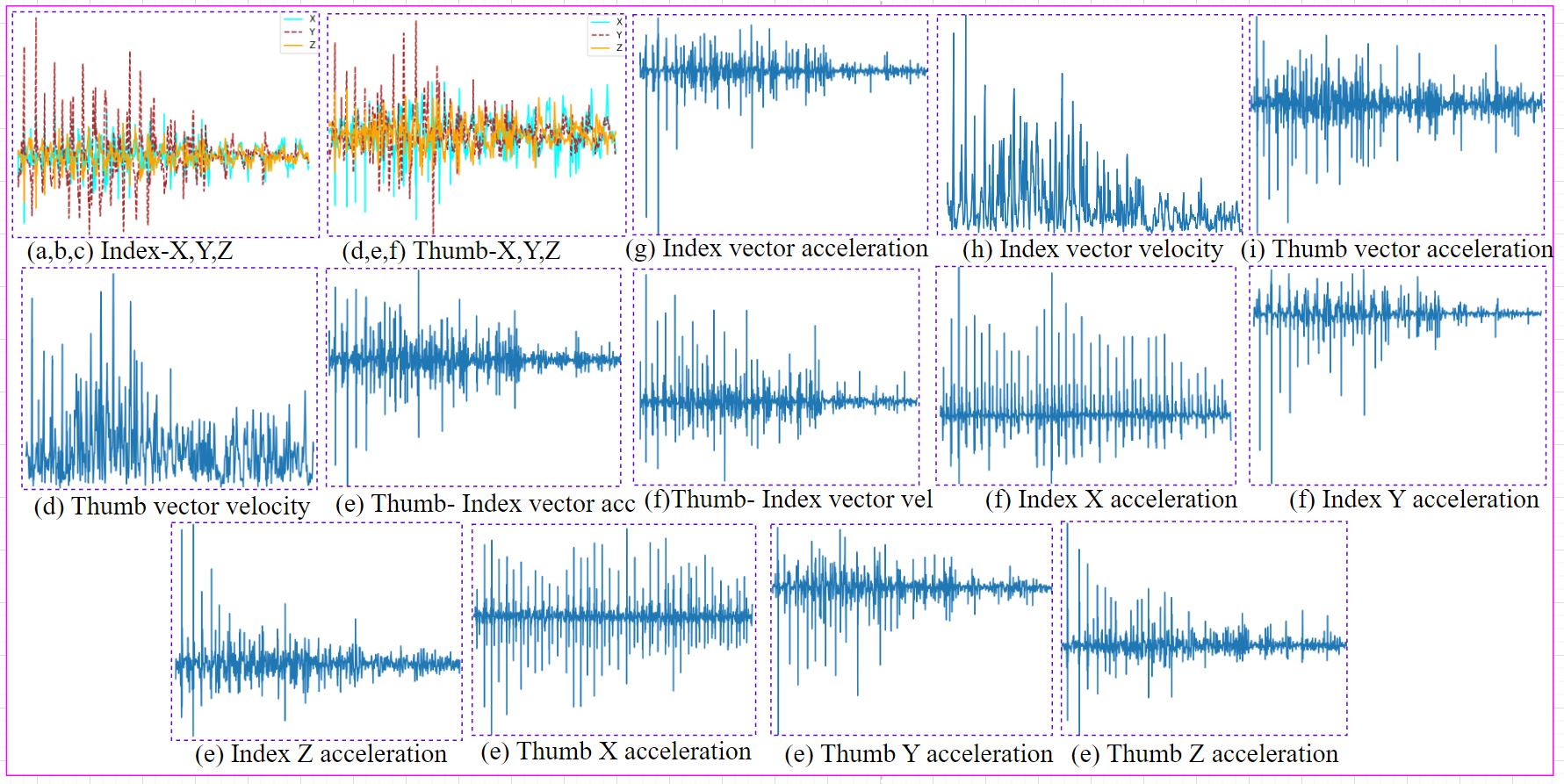}
\caption{\textcolor{red}{Kinematic 18 feature for PD labels.}}
\label{Figure:5}
\end{figure*}

\subsection{Kinematic Feature Extraction }
In the study, we considered 18 kinematic features, where 6 attributes were taken from the input signal X, Y, and Z for both the thumb and index finger. The rest 12 kinematic feature we derived from numerical differentiation and vector magnitude. 
In our study, we analyzed the data using not only separately processed and vectorized signals for each axis but also the relative motion of the thumb and index finger. These signals are divided into three main types of signals: A) Angular velocity (\textit{N = 6}), B) Angular acceleration (\textit{N = 6}), C) Vector (\textit{N = 6})
These features include angular velocities, angular accelerations, vectors for both the thumb and index finger and the relative values between them. The Kinematic feature based 18 signals as depicted in Table \ref{tab:kinamatic_feature} and these signals of HC and PD patients are shown in Figure \ref{Figure:4} and \ref{Figure:5}.
\begin{table*}[]
\centering
\caption{Names and descriptions of the transformed signals \label {tab:kinamatic_feature}}

\begin{tabular}{ll}
\hline
Name of movement pattern  & Description    \\
\hline
Thumb\_\{X,Y,Z\}\_vel & Angular velocity in \{x,y,z\} axes of thumb (Raw Signal). \\ 
Index\_\{X,Y,Z\}\_vel    & Angular velocity in \{x,y,z\} axes of index finger (Raw Signal).        \\
\{Thumb,Index\}\_vec\_vel & Vector of angular velocity of {thumb, index finger}.  \\
Thumb2Index\_vec\_vel & Relative angular velocity vector of thumb and index finger.   \\
Thumb\_\{X,Y,Z\}\_acc  & Angular acceleration in \{x,y,z\} axes of thumb.   \\
Index\_\{X,Y,Z\}\_acc  & Angular acceleration in \{x,y,z\} axes of index finger.   \\
\{Thumb,Index\}\_vec\_acc & Vector of angular acceleration of {thumb, index finger}.  \\
Thumb2Index\_vec\_acc & Relative angular acceleration vector of thumb and index finger.    \\
\hline
\end{tabular}
\end{table*}

\subsubsection{Existing Kinematic Feature}
\paragraph{Angular Velocity of Thumb and Index Finger}
The angular velocity for the thumb along the \(x\), \(y\), and \(z\) axes are computed as the rate of change of the thumb's position (using numerical differentiation):
\begin{equation}
\omega_{\text{Thumb\_X}} = \frac{d\theta_{\text{Thumb\_X}}}{dt} \approx \frac{\text{Thumb\_X}[i+1] - \text{Thumb\_X}[i]}{\Delta t} \label{eq:thumb_x_vel}
\end{equation}

\begin{equation}
\omega_{\text{Thumb\_Y}} = \frac{d\theta_{\text{Thumb\_Y}}}{dt} \approx \frac{\text{Thumb\_Y}[i+1] - \text{Thumb\_Y}[i]}{\Delta t} \label{eq:thumb_y_vel}
\end{equation}
\begin{equation}
\omega_{\text{Thumb\_Z}} = \frac{d\theta_{\text{Thumb\_Z}}}{dt} \approx \frac{\text{Thumb\_Z}[i+1] - \text{Thumb\_Z}[i]}{\Delta t} \label{eq:thumb_z_vel}
\end{equation}

Similarly, for the index finger:
\begin{equation}
\omega_{\text{Index\_X}} = \frac{d\theta_{\text{Index\_X}}}{dt} \approx \frac{\text{Index\_X}[i+1] - \text{Index\_X}[i]}{\Delta t} \label{eq:index_x_vel}
\end{equation}
\begin{equation}
\omega_{\text{Index\_Y}} = \frac{d\theta_{\text{Index\_Y}}}{dt} \approx \frac{\text{Index\_Y}[i+1] - \text{Index\_Y}[i]}{\Delta t} \label{eq:index_y_vel}
\end{equation}
\begin{equation}
\omega_{\text{Index\_Z}} = \frac{d\theta_{\text{Index\_Z}}}{dt} \approx \frac{\text{Index\_Z}[i+1] - \text{Index\_Z}[i]}{\Delta t} \label{eq:index_z_vel}
\end{equation}


\paragraph{Vector of Angular Velocity for Thumb and Index Finger}
The angular velocity vector for the thumb and index finger is computed as the magnitude of the angular velocity in 3D space:
\begin{equation}
\vec{\omega}_{\text{Thumb}} = \sqrt{ \omega_{\text{Thumb\_X}}^2 + \omega_{\text{Thumb\_Y}}^2 + \omega_{\text{Thumb\_Z}}^2 } \label{eq:thumb_vec_vel}
\end{equation}
\begin{equation}
\vec{\omega}_{\text{Index}} = \sqrt{ \omega_{\text{Index\_X}}^2 + \omega_{\text{Index\_Y}}^2 + \omega_{\text{Index\_Z}}^2 } \label{eq:index_vec_vel}
\end{equation}


\paragraph{Angular Acceleration of Thumb and Index Finger}
Angular acceleration is the rate of change of angular velocity. Using numerical differentiation (calculated using the `np.gradient` method), the angular acceleration of the thumb can be written as:
\begin{equation}
\alpha_{\text{Thumb\_X}} = \frac{d\omega_{\text{Thumb\_X}}}{dt} \approx \frac{\omega_{\text{Thumb\_X}}[i+1] - \omega_{\text{Thumb\_X}}[i]}{\Delta t} \label{eq:thumb_x_acc}
\end{equation}

\begin{equation}
\alpha_{\text{Thumb\_Y}} = \frac{d\omega_{\text{Thumb\_Y}}}{dt} \approx \frac{\omega_{\text{Thumb\_Y}}[i+1] - \omega_{\text{Thumb\_Y}}[i]}{\Delta t} \label{eq:thumb_y_acc}
\end{equation}

\begin{equation}
\alpha_{\text{Thumb\_Z}} = \frac{d\omega_{\text{Thumb\_Z}}}{dt} \approx \frac{\omega_{\text{Thumb\_Z}}[i+1] - \omega_{\text{Thumb\_Z}}[i]}{\Delta t} \label{eq:thumb_z_acc}
\end{equation}
Similarly, for the index finger:
\begin{equation}
\alpha_{\text{Index\_X}} = \frac{d\omega_{\text{Index\_X}}}{dt} \approx \frac{\omega_{\text{Index\_X}}[i+1] - \omega_{\text{Index\_X}}[i]}{\Delta t} \label{eq:index_x_acc}
\end{equation}
\begin{equation}
\alpha_{\text{Index\_Y}} = \frac{d\omega_{\text{Index\_Y}}}{dt} \approx \frac{\omega_{\text{Index\_Y}}[i+1] - \omega_{\text{Index\_Y}}[i]}{\Delta t} \label{eq:index_y_acc}
\end{equation}
\begin{equation}
\alpha_{\text{Index\_Z}} = \frac{d\omega_{\text{Index\_Z}}}{dt} \approx \frac{\omega_{\text{Index\_Z}}[i+1] - \omega_{\text{Index\_Z}}[i]}{\Delta t} \label{eq:index_z_acc}
\end{equation}
\paragraph{Vector of Angular Acceleration for Thumb and Index Finger}
The vector of angular acceleration for both the thumb and index finger is computed as the magnitude of their respective angular accelerations in 3D space:
\begin{equation}
\vec{\alpha}_{\text{Thumb}} = \sqrt{ \alpha_{\text{Thumb\_X}}^2 + \alpha_{\text{Thumb\_Y}}^2 + \alpha_{\text{Thumb\_Z}}^2 } \label{eq:thumb_vec_acc}
\end{equation}

\begin{equation}
\vec{\alpha}_{\text{Index}} = \sqrt{ \alpha_{\text{Index\_X}}^2 + \alpha_{\text{Index\_Y}}^2 + \alpha_{\text{Index\_Z}}^2 } \label{eq:index_vec_acc}
\end{equation}

\paragraph{Timestamp Calculation}
Finally, we compute the timestamp for each data point based on the sample rate of 200Hz (sampling time \( \Delta t = 0.005 \) seconds):
\begin{equation}
\text{Timestamp}[i] = i \times 0.005 \quad \text{where} \quad i \in \{0, 1, 2, \dots, n-1\} \label{eq:timestamp}
\end{equation}
\subsubsection{Newly Proposed Kinematic Feature}
In this study, we introduce two newly proposed kinematic features: Thumb to Index vector velocity and acceleration. These features represent the relative angular velocity and acceleration between the thumb and index finger, respectively. The advantage of these features lies in their ability to capture the dynamic interaction between two critical fingers involved in motor tasks. Unlike traditional single-axis or single-finger measurements, these features offer a more holistic representation of movement, which is crucial for differentiating between various motor symptoms in neurodegenerative disorders such as Parkinson’s disease (PD), Multiple System Atrophy (MSA) and Progressive Supranuclear Palsy (PSP). By incorporating these relative movement patterns, sensor-based systems can achieve higher sensitivity and specificity in recognizing subtle motor impairments. This enhancement can significantly improve the accuracy of PD recognition systems, making them more reliable for clinical diagnosis and monitoring. The calculation procedure for these two features is given below: 

\paragraph{Relative Angular Velocity between Thumb and Index Finger}
The relative angular velocity vector between the thumb and index finger is calculated as the difference between their individual velocity vectors:

\begin{equation}
\omega_{\text{Thumb2Index\_X}} = \omega_{\text{Thumb\_X}} - \omega_{\text{Index\_X}}
\label{eq:thumb2index_X}
\end{equation}

\begin{equation}
\omega_{\text{Thumb2Index\_Y}} = \omega_{\text{Thumb\_Y}} - \omega_{\text{Index\_Y}}
\label{eq:thumb2index_Y}
\end{equation}

\begin{equation}
\omega_{\text{Thumb2Index\_Z}} = \omega_{\text{Thumb\_Z}} - \omega_{\text{Index\_Z}}
\label{eq:thumb2index_Z}
\end{equation}

\begin{equation}
\vec{\omega}_{\text{Thumb2Index}} = \sqrt{ 
\omega_{\text{Thumb2Index\_X}}^2 + \omega_{\text{Thumb2Index\_Y}}^2 + \omega_{\text{Thumb2Index\_Z}}^2}
\label{eq:thumb2index_vec_vel}
\end{equation}

\paragraph{Relative Angular Acceleration between Thumb and Index Finger}
The relative angular acceleration vector between the thumb and index finger is calculated as the difference between their angular acceleration vectors:
\begin{equation}
\alpha_{\text{Thumb2Index\_X}} = \alpha_{\text{Thumb\_X}} - \alpha_{\text{Index\_X}}
\label{eq:thumb2index_X_acc}
\end{equation}

\begin{equation}
\alpha_{\text{Thumb2Index\_Y}} = \alpha_{\text{Thumb\_Y}} - \alpha_{\text{Index\_Y}}
\label{eq:thumb2index_Y_acc}
\end{equation}

\begin{equation}
\alpha_{\text{Thumb2Index\_Z}} = \alpha_{\text{Thumb\_Z}} - \alpha_{\text{Index\_Z}}
\label{eq:thumb2index_Z_acc}
\end{equation}

\begin{equation}
\vec{\alpha}_{\text{Thumb2Index}} = \sqrt{ 
\alpha_{\text{Thumb2Index\_X}}^2 + \alpha_{\text{Thumb2Index\_Y}}^2 + \alpha_{\text{Thumb2Index\_Z}}^2}
\label{eq:thumb2index_vec_acc}
\end{equation}
\subsection{Statistical Feature extractin}
In this study, 27 new features were calculated in addition to the 14 features used in the previous study. Finally, 41 features were extracted from each signal, for a total of 738 features (18 signals x 41 features). Table \ref{tab:Table 3} shows the 41 extracted features obtained from each signal.


\subsubsection{Existing Statistical Feature}
Following previous study\cite{belic2023quick}, 6 features were extracted from the entire signal and the local maxima of the signal. We used the AMPD (automatic multiscale-based peak detection) algorithm to identify local maxima\cite{scholkmann2012efficient}. In addition, each signal was transformed by Fourier, and the maximal frequency and the spectral centroid were extracted. Therefore, a total of 14 features were calculated. 

\begin{table*}
\centering
\caption{List of extracted features from raw features  \label {tab:Table 3}}

\resizebox{\textwidth}{!}{

\begin{tabular}{l|l|l} 

\textbf{Features Names }  & \textbf{Explanations } & \textbf{Formula} \\ \hline
RMS \cite{belic2023quick}  & Root mean square of signal & $\sqrt{\frac{1}{N} \sum_{i=1}^{N} X_i^2}$  \\ 
\hline
Min \cite{belic2023quick}  &Minimum value of signal & $\textit{Min}(X)$  \\  \hline

Max \cite{belic2023quick}  &Maximum value of signal & $\textit{Max}(X)$  \\  \hline

Avg \cite{belic2023quick}  &Average value of signal & $\frac{1}{N} \sum_{i=1}^{N} X_i$  \\  \hline

Std \cite{belic2023quick}  &Standard deviation of signal & $\sqrt{\frac{1}{N} \sum_{i=1}^{N} (X_i - \bar{X})^2}$  \\  \hline

Med \cite{belic2023quick}  &Median of signal & $\textit{Med}(X)$  \\  \hline

Max\_freq \cite{belic2023quick}  &Maximal frequency resulting from the Fourier transform & $\textit{Max}(\omega)$  \\  \hline

Centroid \cite{belic2023quick}  &Spectral centroid resulting from the Fourier transform & $\frac{\sum_{i=1}^{n} \omega_i P(\omega_i)}{\sum_{i=1}^{n} P(\omega_i)}$  \\  \hline

RMS\_Max \cite{belic2023quick}  & Root mean square of maximum points & $\sqrt{\frac{1}{M} \sum_{i=1}^{M} MP_i^2}$  \\  \hline

Min\_Max \cite{belic2023quick}  &Minimum value of maximum points& $\textit{Min}(MP)$  \\  \hline

Max\_Max \cite{belic2023quick}  &Maximum value of maximum points& $\textit{Max}(MP)$  \\  \hline

Avg\_Max \cite{belic2023quick}  &Average value of maximum points & $\frac{1}{M} \sum_{i=1}^{M} MP_i$  \\  \hline

Std\_Max \cite{belic2023quick}  &Standard deviation of maximum points & $\sqrt{\frac{1}{M} \sum_{i=1}^{M} (MP_i - \bar{MP})^2}$  \\  \hline

Med\_Max \cite{belic2023quick}  &Median of maximum points & $\textit{Med}(MP)$  \\  \hline

Noise\_Var \cite{garcia2010robust}   & Noise variance &   \cite{garcia2010robust}  \\  \hline

Conv\_Ene \cite{drotar2014decision} & Amplitude sum of squares & $\sum_{i=1}^{N}{X_i}^2 $    \\  \hline

SNR \cite{drotar2014decision}    & Signal to noise ratios & $\ Conv\_Ene/Noise\_Var $    \\  \hline

Var \cite{christ2018time}  & Variance of time-series amplitude d & $\frac{1}{N}\sum_{i=1}^{N}\left(X_i-\ \bar{X}\right)^2 $  \\  \hline

Avg\_Abs\_Change \cite{christ2018time}   & Average over first amplitude differences & $\frac{1}{N-1}\sum_{i=1}^{N-1}\left|X_{i+1}-X_i\right| $    \\  \hline

Autocorr\_Lagi ; i=1,…,9 \cite{christ2018time} & Autocorrelation of the specified lag  & \cite{christ2018time}  \\  \hline

Quant\_i ; i=0.1,0.2…0.9 (excludeing 0.5) \cite{christ2018time}& Quantile of time-series velocity & \cite{christ2018time}  \\
\hline

Rhythm & Regularity of signal oscillations & Frequency domain analysis \\\hline

Amplitude & Peak-to-peak value of signal & $\textit{Max}(X) - \textit{Min}(X)$ \\ \hline
Frequency & Dominant frequency component of the signal & $\textit{argmax}_\omega \left( P(\omega) \right)$ \\ \hline
Frequency\_Std & Standard deviation of frequency components & $\sqrt{\frac{1}{n} \sum_{i=1}^{n} (\omega_i - \bar{\omega})^2}$ \\
\hline
Slope & Rate of change of the signal & $\frac{X_{N} - X_{1}}{N}$
\\ \hline

\end{tabular}}

\footnotesize Here, N: length of data. M: Number of extrema. P($\omega_i$):  the spectral power at frequency $\omega_i$. MPi: ith value of the local maximum values of a signal;
\end{table*}

\begin{enumerate}
\item RMS (Root Mean Square): Measures the average magnitude of the signal by computing the square root of the mean of the squared values. The formula is shown in Equation~\ref{eq:rms}:
\begin{equation}
\text{RMS} = \sqrt{\frac{1}{N} \sum_{i=1}^{N} X_i^2}
\label{eq:rms}
\end{equation}

\item Min (Minimum Value): The smallest value in the signal as shown in Equation~\ref{eq:min}:

\begin{equation}
\textit{Min}(X)
\label{eq:min}
\end{equation}

\item Max (Maximum Value): The largest value in the signal as shown in Equation~\ref{eq:max}:

\begin{equation}
\textit{Max}(X)
\label{eq:max}
\end{equation}

\item Avg (Average Value): The mean of all the signal values as shown in Equation~\ref{eq:avg}:

\begin{equation}
\text{Avg}(X) = \frac{1}{N} \sum_{i=1}^{N} X_i
\label{eq:avg}
\end{equation}

\item Std (Standard Deviation): Measures the amount of variation or dispersion in the signal as shown in Equation~\ref{eq:std}:

\begin{equation}
\text{Std}(X) = \sqrt{\frac{1}{N} \sum_{i=1}^{N} (X_i - \bar{X})^2}
\label{eq:std}
\end{equation}

\item Med (Median): The middle value of the signal when sorted in ascending order as shown in Equation~\ref{eq:med}:

\begin{equation}
\textit{Med}(X)
\label{eq:med}
\end{equation}

\item Max\_freq (Maximum Frequency): The highest frequency present in the signal after applying the Fourier transform as shown in Equation~\ref{eq:max_freq}:

\begin{equation}
\textit{Max}(\omega)
\label{eq:max_freq}
\end{equation}

\item Centroid (Spectral Centroid): The center of mass of the power spectrum of the signal as shown in Equation~\ref{eq:centroid}:

\begin{equation}
\text{Centroid} = \frac{\sum_{i=1}^{n} \omega_i P(\omega_i)}{\sum_{i=1}^{n} P(\omega_i)}
\label{eq:centroid}
\end{equation}

\item RMS\_Max (RMS of Maximum Points): The RMS value computed only over the local maximum points of the signal as shown in Equation~\ref{eq:rms_max}:

\begin{equation}
\text{RMS\_Max} = \sqrt{\frac{1}{M} \sum_{i=1}^{M} MP_i^2}
\label{eq:rms_max}
\end{equation}

\item Min\_Max (Minimum of Maximum Points): The smallest value among the local maxima of the signal as shown in Equation~\ref{eq:min_max}:

\begin{equation}
\textit{Min}(MP)
\label{eq:min_max}
\end{equation}

\item Max\_Max (Maximum of Maximum Points): The largest value among the local maxima of the signal as shown in Equation~\ref{eq:max_max}:

\begin{equation}
\textit{Max}(MP)
\label{eq:max_max}
\end{equation}

\item Avg\_Max (Average of Maximum Points): The mean of the local maxima of the signal as shown in Equation~\ref{eq:avg_max}:

\begin{equation}
\text{Avg\_Max}(MP) = \frac{1}{M} \sum_{i=1}^{M} MP_i
\label{eq:avg_max}
\end{equation}

\item Std\_Max (Standard Deviation of Maximum Points): The standard deviation of the local maxima of the signal as shown in Equation~\ref{eq:std_max}:

\begin{equation}
\text{Std\_Max}(MP) = \sqrt{\frac{1}{M} \sum_{i=1}^{M} (MP_i - \bar{MP})^2}
\label{eq:std_max}
\end{equation}

\item Med\_Max (Median of Maximum Points): The median value of the local maxima of the signal as shown in Equation~\ref{eq:med_max}:
\begin{equation}
\textit{Med}(MP)
\label{eq:med_max}
\end{equation}
\end{enumerate}
\subsubsection{Newly Proposed Statistical Feature}
Besides the existing statistical feature, we also propose a new statistical feature: In addition to the kinematic features, we propose several new statistical features that contribute significantly to the effectiveness of sensor-based systems for Parkinson’s Disease (PD) recognition. Among these, the Signal-to-Noise Ratio (SNR) is calculated as the ratio of Convolved Energy  to Noise Variance. SNR is an essential metric that quantifies the signal's strength relative to the noise, providing insight into the clarity of the captured data. A higher SNR indicates better-quality signals, which are crucial for differentiating meaningful motor activity from noise, especially in real-world environments where sensor noise is prevalent.

Other statistical features include Variance (Var), which measures the spread of signal values, and Average Absolute Change, which quantifies the signal’s temporal variability. Rhythm captures the regularity of the signal’s oscillations, while Amplitude measures the peak-to-peak signal magnitude, reflecting the strength of the movement. Frequency identifies the dominant frequency component, providing insight into periodicity, and Standard Deviation of Frequency gauges the spread of frequency components. Lastly, the Slope feature indicates the rate of change of the signal, revealing rapid transitions or movements. These features collectively enhance the feature set by capturing both time-domain and frequency-domain characteristics of the signal. By incorporating these statistical features alongside traditional kinematic features, we improve the robustness and accuracy of the classification models. This enables the system to more effectively classify PD, PSP, MSA, and HC, even in the presence of noise or other interference. The integration of these features strengthens the system's reliability, making it a more stable and effective diagnostic tool for movement disorders in clinical settings.

\begin{enumerate}
\item Conv\_Ene (Convolved Energy): The sum of the squared amplitudes of the signal as shown in Equation~\ref{eq:conv_ene}:
\begin{equation}
\text{Conv\_Ene} = \sum_{i=1}^{N} X_i^2
\label{eq:conv_ene}
\end{equation}
\item SNR (Signal to Noise Ratio): The ratio of the signal's energy to its noise variance as shown in Equation~\ref{eq:snr}:
\begin{equation}
\text{SNR} = \frac{\text{Conv\_Ene}}{\text{Noise\_Var}}
\label{eq:snr}
\end{equation}
\item Var (Variance): The variance of the signal's amplitude as shown in Equation~\ref{eq:var}:
\begin{equation}
\text{Var}(X) = \frac{1}{N} \sum_{i=1}^{N} (X_i - \bar{X})^2
\label{eq:var}
\end{equation}
\item Avg\_Abs\_Change (Average Absolute Change): The mean of the absolute differences between consecutive signal values as shown in Equation~\ref{eq:avg_abs_change}:
\begin{equation}
\text{Avg\_Abs\_Change} = \frac{1}{N-1} \sum_{i=1}^{N-1} \left| X_{i+1} - X_i \right|
\label{eq:avg_abs_change}
\end{equation}
\item Rhythm: Refers to the regularity of the signal’s pattern or oscillation over time. It is computed by analyzing the periodic components of the signal through a Fourier transform. While rhythm doesn’t have a specific formula, it's typically characterized by the signal’s frequency domain representation.
\item Amplitude: The peak-to-peak value of the signal, which measures the difference between the maximum and minimum values of the signal. It reflects the signal's magnitude as shown in Equation~\ref{eq:amplitude}:

\begin{equation}
\text{Amplitude}(X) = \textit{Max}(X) - \textit{Min}(X)
\label{eq:amplitude}
\end{equation}

\item Frequency: The dominant frequency component of the signal, typically obtained using the Fourier transform. It is the frequency with the highest magnitude in the power spectrum, as shown in Equation~\ref{eq:frequency}:

\begin{equation}
\text{Frequency}(X) = \textit{argmax}_{\omega} \left( P(\omega) \right)
\label{eq:frequency}
\end{equation}

\item Frequency\_Std (Standard Deviation of Frequency): The standard deviation of the signal's frequency components, which reflects how spread out the frequencies are around the dominant frequency, as shown in Equation~\ref{eq:frequency_std}:

\begin{equation}
\text{Frequency\_Std}(X) = \sqrt{\frac{1}{n} \sum_{i=1}^{n} (\omega_i - \bar{\omega})^2}
\label{eq:frequency_std}
\end{equation}

\item Slope: The rate of change of the signal over time. It indicates the steepness of the signal's progression and can be computed using the first derivative, as shown in Equation~\ref{eq:slope}:

\begin{equation}
\text{Slope}(X) = \frac{X_{N} - X_{1}}{N}
\label{eq:slope}
\end{equation}

\end{enumerate}

\subsection{Feature Selection}

In this study, 738 features were extracted for a single data. However, some of these features do not contribute to or hinder classification.  In this study, two feature selection methods (One-way ANOVA and SFFS) were performed to identify features important for differential diagnosis. We initially employed a One-way ANOVA to order the input features based on \textit{p}-value. We computed the \textit{p}-value for each feature and only used features with a \textit{p}-value smaller than 0.005 for classification. We then combined SFFS with SVM to find appropriate feature combinations using features selected by One-way ANOVA.

\subsubsection{One-way ANOVA}
One-way ANOVA (Analysis of Variance) is a statistical method used to determine whether there is a statistically significant difference between the means of three or more independent groups. In this study, One-way ANOVA was applied to identify the features that exhibited significant differences among the four groups: HC, PD, PSP, and MSA\cite{miah2020motor_oneway_anova,miah2017motor}. 
The hypothesis tested in One-way ANOVA is:

$H_0: \mu_1 = \mu_2 = \dots = \mu_k \\
\quad \text{(Null Hypothesis: All group means are equal)}$ \\
$H_1: \text{At least one group mean is different} \\
\quad \text{(Alternative Hypothesis: At least one group mean differs)} \\
\label{eq:alt_hypothesis}$

Here, \( \mu_i \) represents the mean of the i-th group, and \( k \) is the number of groups.
The F-statistic used in One-way ANOVA is calculated as:
\begin{equation}
F = \frac{\text{Between-group variance}}{\text{Within-group variance}} \\
= \frac{\frac{1}{k-1} \sum_{i=1}^{k} n_i (\bar{X}_i - \bar{X})^2}{\frac{1}{N-k} \sum_{i=1}^{k} \sum_{j=1}^{n_i} (X_{ij} - \bar{X}_i)^2}
\label{eq:f_statistic}
\end{equation}

Where:\\
- \( n_i \) is the number of observations in group \( i \),\\
- \( \bar{X}_i \) is the mean of group \( i \),\\
- \( \bar{X} \) is the overall mean of all groups,\\
- \( N \) is the total number of observations across all groups.\\

The \textit{p}-value for each feature is then calculated, and features with \textit{p}-values less than 0.005 are considered significant. The \textit{p}-value is determined by the F-distribution:
\begin{equation}
p = P(F > f_{\text{observed}}) \quad 
\label{eq:p_value}
\end{equation}
Here, \textit{p}-value calculation is based on the F-distribution. Once the \textit{p}-values are computed for each feature, the features with the smallest \textit{p}-values are selected. Features with \( p < 0.005 \) are considered significant for further analysis. 

To reduce dimensionality, Principal Component Analysis (PCA) is applied, and the features are ranked based on their \textit{p}-values. The components selected by PCA are then ranked, and the minimum number of components is determined using the weighted False Discovery Rate (FDR), with the expected ratio of false rejections among the rejected hypotheses. 

The FDR-adjusted \textit{p}-values are calculated using the formula:

\begin{equation}
p_{\text{adj}} = \frac{p \cdot k}{i}
\label{eq:fdr_adj}
\end{equation}

Where: \\
- \( p \) is the raw \textit{p}-value, \\
- \( k \) is the total number of tests (features),\\
- \( i \) is the rank of the \textit{p}-value in ascending order.\\

We determined a threshold of \( p_{\text{adj}} < 0.8 \) based on the FDR function to select the efficient features. The selected features were then used for further analysis.

Finally, the selected features are transformed using the PCA equation:

\begin{equation}
S_0 = U^T S
\label{eq:pca_transformation}
\end{equation}

Where: \\
- \( S_0 \) is the transformed feature vector,\\
- \( U \) is the matrix of eigenvectors obtained from PCA,\\
- \( S \) is the matrix of feature values.\\

This procedure ensures that the most relevant features are selected for further analysis while controlling for the false discovery rate in the feature selection process.

\subsubsection{SFFS-Based Algorithm}
SFFS is a widely used feature selection algorithm that plays an important role in enhancing generalization errors in machine learning. This algorithm can find the optimal feature set very efficiently because it can add features sequentially while excluding unnecessary features as needed. The detailed steps of the SFFS-based algorithm are comprehensively documented elsewhere \cite{pudil1994floating}. In this study, SFFS was used in combination with SVM.

\subsection{Classification}
We implemented several machine learning methods, including k-Nearest Neighbors (kNN) and SVM. SVM is a powerful algorithm for classification and regression tasks, particularly effective for high-dimensional data. By using kernel functions, SVM can excel in classifying both linear and nonlinear data. In this study, we used an SVM to classify patients into four categories (HC, PD, PSP, and MSA). In addition, to evaluate the performance of our proposed method, we employed kNN as a baseline and compared it with a classifier using the same feature selection method. In this study, we calculate the classification accuracy for each subject and each data. Predictions are obtained for all recorded data, and each patient's predicted diagnosis takes the most frequently predicted diagnosis in their records as the patient's predicted diagnosis. However, if there are the predicted diagnosis with the same frequency in their records, the more probable diagnosis given by the probability score was adopted.


\section{Experimental Results}
\label {sec: Experimental Setup and Performance Metrics}
We calculated 18 different signals for each data set. Six of these signals represented angular velocities, six signals differentiated the angular velocities and represented angular acceleration, and the remaining six represented vectors. 41 Statistical features were extracted from each of these signals. The formulas for these statistical features are shown in Table \ref{tab:Table 3}. This comprehensive process yielded 738 features (41 features x 18 signals), which were used in both One-way ANOVA and SFFS-based feature selection to identify the most-relevant features for detecting PD and atypical parkinsonism.
After extracting 738 features, we applied the One-way ANOVA algorithm to compute the \textit{p}-value of each feature. We used only the features with \textit{p}-values less than 0.005, and finally, we selected 534 features. These 534 features were used in the SFFS algorithm integrated with the SVM model to search for the optimal combination of features that can accurately classify diseases. In this study, the SVM model was trained using LOOCV.

\subsection{Experimental Setup and Performance Matrix}
All experiments were performed on a Windows 11 computer. The data used for analysis were gyroscope data available on GitHub and were analyzed using Python (version 3.10.13). SVM was used for classification, with LOOCV.

To predict diseases, we trained an SVM model and constructed a confusion matrix to compare the predicted classes with the actual classes. As mentioned above, predictions were made for all recorded data, and each subject's predicted diagnosis was determined by the most frequent predicted diagnosis among the subject's records. Therefore, we created confusion matrices for both individual data points and for each subject. Additionally, we calculated four performance metrics (accuracy, recall, precision, and F1-score) from these confusion matrices to evaluate the performance of the SVM. Their formulae have already been clearly explained elsewhere 
\cite{miah2022bensignnet_performance_matrix,miah2021alzheimer_disease,
ali2022potential_miah,kafi2022lite_kidney_miah}.

SVM hyperparameters were fine-tuned using Optuna. During the training phase, we systematically explored two kernel functions (RBF and sigmoid) and varied the cost and $\gamma$ values within the range of 0.01 to 100. We conducted 200 trials with a fixed seed value of 42 to optimize the parameters. After optimization, we retrained the SFFS with the SVM model for (\textit{N}-1) PD patients, reserving one patient as the test set for prediction evaluation. This procedure was repeated \textit{N} times.

\subsection{Ablation Study}
Table \ref{tab:Table 5} presents an ablation study comparing the performance of two machine learning algorithms, SVM and kNN, using varying sets of features to assess their impact on model accuracy.
For the SVM model, the accuracy achieved is 88.89\%, with 10 selected features. Of these, five are newly proposed and highlighted in bold which are emphasizing their significance in improving classification performance.

The kNN model, with a slightly lower accuracy of 87.04\%, uses 9 features, 7 of which are new and also marked in bold, demonstrating the strong impact of the new features on model performance.

\begin{table*}[]
\caption{Ablation study of the proposed study with various machine learning algorithms. \label {tab:Table 5}}
\begin{tabular}{|l|l|l|l|l|p{5cm}|}
\hline
Algorithm Name & Accuracy & Total selected features & Features Name & \textit{p}-value & Our proposed new features in selected features \\ \hline

\multirow{10}{*}{SVM} & \multirow{10}{*}{88.89} & \multirow{10}{*}{10} & \textbf{Thumb\_vec\_vel\_quantile\_q\_0.4} & 5.73e-27 & \multirow{10}{*}{5 (\textbf{Bold in Features name column})} \\
 &  &  & Thumb\_x\_vel\_std\_of\_max\_taps & 1.96e-12 &  \\
 &  &  & \textbf{Index\_x\_acc\_freqstd} & 9.11e-12 &  \\
 &  &  & \textbf{Thumb\_y\_acc\_SNR} & 1.84e-05 &  \\
 &  &  & Thumb\_x\_vel\_std & 4.21e-07 &  \\
 &  &  & \textbf{Thumb\_y\_vel\_variance} & 8.29e-08 &  \\
 &  &  & Thumb\_x\_acc\_avg\_of\_max\_taps & 1.02e-18 &  \\
 &  &  & Thumb\_z\_acc\_median & 6.04e-08 &  \\
 &  &  & \textbf{Thumb2Index\_vec\_vel\_noise\_var} & 7.42e-06 &  \\
 &  &  & Thumb\_vec\_acc\_RMS\_of\_max\_taps & 2.71e-08 &  \\
 \hline

 \multirow{9}{*}{kNN} & \multirow{9}{*}{87.04} & \multirow{9}{*}{9} & 
 \textbf{Index\_z\_acc\_autocorrelation\_lag\_3} & 7.56e-11 & \multirow{9}{*}{7 (\textbf{Bold in Features name column})} \\
 &  &  & \textbf{Thumb\_y\_vel\_autocorrelation\_lag\_8} & 2.75e-6 &  \\
 &  &  & \textbf{Thumb\_z\_acc\_autocorrelation\_lag\_4} & 1.36e-5 &  \\
 &  &  & \textbf{Index\_y\_acc\_quantile\_q\_0.6} & 7.87e-46 &  \\
 &  &  & \textbf{Thumb\_z\_acc\_quantile\_q\_0.4} & 1.41e-19 &  \\
 &  &  & \textbf{Index\_vec\_vel\_quantile\_q\_0.1} & 1.95e-38 &  \\
 &  &  & Index\_y\_acc\_median & 4.35e-22 &  \\
 &  &  & \textbf{Thumb\_vec\_acc\_quantile\_q\_0.6} & 1.31e-24 &  \\
 &  &  & Thumb\_y\_acc\_max\_freq & 1.51e-3 &  \\
 
 \hline
\end{tabular}
\end{table*}

\subsection{Performance Result of the Proposed Model}
The \textit{p}-values for each feature are shown in Table \ref{tab:Table 5}. To assess the performance of each feature, we used SVM algorithm. In this study, we primarily focused on selecting the most important features using the Sequential Forward Floating Selection (SFFS) algorithm. After selecting the top features, we fed them into the SVM algorithm and observed that the highest performance in the FT task was achieved with a combination of 10 selected features.

Table \ref{tab:performance_result} presents the performance accuracy of the proposed model, where various metrics such as classification accuracy, recall, precision, and F1-score were calculated based on the confusion matrices. The classification accuracy for each subject was 66.67\%, while for each data set, it was 88.89\%.

In addition to the overall results, we also computed the classification accuracy for each data set and subject individually. The predictive label for each subject was determined by the most frequently predicted label from the data collected for that subject. In cases where the predicted diagnoses appeared with equal frequency, the diagnosis with the highest SVM probability score was selected as the final prediction.

\begin{figure}[ht]
\centering
\includegraphics[width=3.5 in]{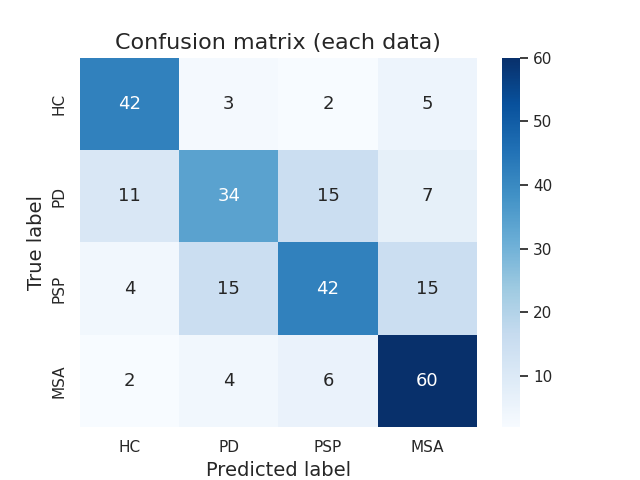}
\caption{Confusion matrix for each data.}
\label{Figure:7}
\end{figure}

\begin{figure}[ht]
\centering
\includegraphics[width=3.5 in]{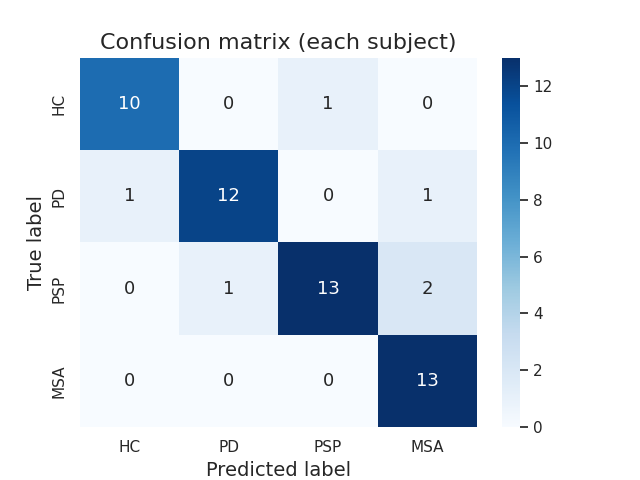}
\caption{Confusion matrix for each subject.}
\label{Figure:8}
\end{figure}

Figures \ref{Figure:7} and \ref{Figure:8} present the confusion matrices comparing the actual and predicted class labels.
According to the confusion matrix shown in Figure \ref{Figure:7} (for each data set), MSA was highly predictive, with 60 cases correctly classified. HC were also classified accurately, with 42 cases correctly predicted. Table \ref{tab:each_class_performance_data} shows the precision, recall, and F1-score for each class. From this table, the results indicate that the model achieved the highest F1-score for HC. Therefore, in our model, healthy subjects are less likely to be misdiagnosed as having other diseases. However, the classification accuracy for PD was relatively low, as it was frequently misclassified as either PSP or HC.
In Figure \ref{Figure:8} (for each subject), the classification accuracy was further improved, reaching 88.89\%. The classification accuracy for HC and MSA was particularly high, with 10 and 13 patients correctly classified, respectively. One healthy subject was misclassified as PSP. This occurred due to a split in the predictions from four data of the same subject, with two predicting HC and two predicting PSP. The final classification was determined by the SVM prediction probabilities, which were slightly higher for PSP and thus classified as PSP. In PD and PSP, 12 and 13 patients were correctly classified, while several patients were incorrectly classified into other categories. In particular, it can be seen that many patients were misclassified in MSA.



\begin{table}[ht]
\centering
\caption{Classification performance (in \%) of the SVM for predicting diseases. \label {tab:performance_result} }
\begin{tabular}{lllll}
\hline
 Target & Accuracy & Precision & Recall & F1-score \\
\hline
 Each data  & 66.67    & 66.37     & 67.53  & 66.50   \\
 Each subject  & 88.89    & 89.33     & 89.47  & 89.03   \\
\hline
\end{tabular}
\end{table}

\begin{table}[ht]
\centering
\caption{Accuracy, precision, recall and F1-score for each class (with each data). \label {tab:each_class_performance_data} }
\begin{tabular}{lllll}
\hline
  & Precision & Recall & F1-score \\
\hline
 HC  & 71.19    & 80.77     & 75.68 \\
 PD  & 60.71    & 50.75    & 55.28 \\
 PSP  & 64.62    & 55.26     & 59.57 \\
 MSA  & 68.97    & 83.33     & 75.47 \\
 \hline
 Accuracy  &     &      & \textbf{66.67} \\
\hline
\end{tabular}
\end{table}

\begin{table}[ht]
\centering
\caption{Accuracy, precision, recall and F1-score for each class (with each subject). \label {tab:each_class_performance_subject} }
\begin{tabular}{lllll}
\hline
  & Precision & Recall & F1-score \\
\hline
 HC  & 90.91    & 90.91     & 90.91 \\
 PD  & 92.31    & 85.71    & 88.89 \\
 PSP  & 92.86    & 81.25     & 86.67 \\
 MSA  & 81.25    & 100.0     & 89.66 \\
 \hline
 Accuracy  &     &      & \textbf{88.89} \\
\hline
\end{tabular}
\end{table}

\begin{table*}[]
\centering
\caption{State of the Art Comparison for the Proposed Model. \label {tab:Table10} }
\begin{tabular}{llllll}
\hline
Authors  & Device   & Task   & Target   \& Class   & Classifier & Accuracy (\%)   \\
\hline
Belic et al. \cite{belic2023quick} & SCU  & FT & \begin{tabular}[c]{@{}l@{}} 4 class \\ (PD, PSP, MSA, and HC)\end{tabular} & kNN  & \begin{tabular}[c]{@{}l@{}}Each data: - \\  Each subject: 85.18\end{tabular} \\

This study  & SCU & FT & \begin{tabular}[c]{@{}l@{}} 4 class \\ (PD, PSP, MSA, and HC)\end{tabular} & SVM  & \begin{tabular}[c]{@{}l@{}}Each data: 66.67\\  Each subject: 88.89\end{tabular} \\
\hline
\end{tabular}
\end{table*}
\subsection{State of the Art Comparison for the Proposed Model}
This section presents a comparative analysis of the performance of our proposed system with previous studies, as summarized in Table \ref{tab:Table10}. Belic et al. \cite{belic2023quick} proposed a differential diagnosis method for patients based on the FT task using wearable sensors.  The existing system follow the four steps. (i) Recording a total of 268 data points from 54 patients performing the FT task using SCU, (ii) Applying five types of transformations to the raw data, (iii) Extracting a total of 216 features from these signals, and (iv) Using a semi-greedy algorithm to select the best features and classifying the data using kNN. As a result, their classification accuracy was 85.18\%. On the other hand, our study also proposes an ML-based differential diagnosis system. In our approach, we analyze the raw data and calculate the relative angular velocity vectors of the thumb and index finger. In addition to the features used in existing studies, we extract features related to signal noise \cite{garcia2010robust,drotar2014decision} and more detailed features using the \texttt{tsfresh} library \cite{christ2018time}. Furthermore, we apply One-way ANOVA to the extracted features, calculate their respective \textit{p}-values, and use the significantly different features in the Sequential Forward Floating Search (SFFS) with Support Vector Machine (SVM) models to identify the most relevant feature combinations. As a result, our approach achieves a classification accuracy of 88.89\%. This demonstrates the superior performance of our system compared to conventional methods.

\section{Conclusion and Future Work}
\label {sec: Conclusion and future plan}
In the study, we proposed a machine learning-based differential diagnosis system employing a set of kinematic feature-based hierarchical feature extraction and selection approaches. The system involves calculating kinematic features from input signals, followed by the extraction of newly proposed statistical features alongside traditional features. By integrating kinematic feature-based extraction and statistical analysis, the system effectively distinguished between these groups, achieving promising classification accuracy. The inclusion of newly proposed features, such as Thumb-to-index vector velocity and acceleration, contributed valuable insights into motor control patterns, while feature selection techniques optimized the model’s efficiency. Feature selection was performed using One-way ANOVA to rank features, and Sequential Forward Floating Selection (SFFS) was used to identify the most relevant features. The resulting hierarchical feature set was then applied to classify PD, PSP, MSA, and healthy controls (HC) using various machine learning classifiers. The system achieved a classification accuracy of 66.67\% for each dataset and 88.89\% for each patient, with notably high performance for the MSA and HC groups using the SVM algorithm. This demonstrates the potential of the proposed system as an accurate and rapid diagnostic support tool in clinical settings.  In the future, we plan to work on PSP classification with more data and refined features, including the gesture-based sensor dataset.
\bibliographystyle{IEEEtran}

\bibliography{Mybib}


\begin{IEEEbiography}[{\includegraphics[width=1in,height=1.25in,clip,keepaspectratio]{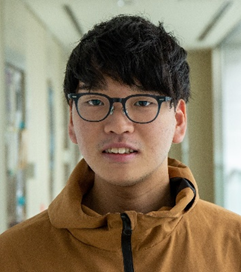}}]{Masahiro Matsumoto} received a bachelor’s degree in computer science and engineering from The University of Aizu (UoA), Japan, in March 2023. He is currently pursuing a master’s degree. He joined the Pattern Processing Laboratoratory, UoA, in April 2022, under the supervision of Prof. Dr. Jungpil Shin. His research interests include computer vision, pattern recognition, and deep learning. He is currently working on human activity recognition, human gesture recognition, and Parkinson’s disease diagnosis. 
\end{IEEEbiography}

\begin{IEEEbiography}    [{\includegraphics[width=1in,height=1.25in,clip,keepaspectratio]{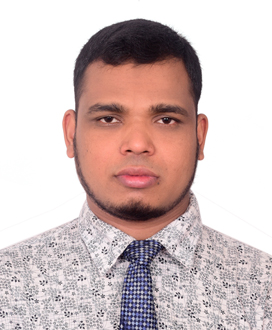}}]
{Abu Saleh Musa Miah} received the B.Sc.Engg. and M.Sc.Engg. degrees in computer science and engineering from the Department of Computer Science and Engineering, University of Rajshahi, Rajshahi-6205, Bangladesh, in 2014 and 2015, respectively, achieving the first merit position. He received his Ph.D. in computer science and engineering from the University of Aizu, Japan, in 2024, under a scholarship from the Japanese government (MEXT). He assumed the positions of Lecturer and Assistant Professor at the Department of Computer Science and Engineering, Bangladesh Army University of Science and Technology (BAUST), Saidpur, Bangladesh, in 2018 and 2021, respectively. Currently, he is working as a visiting researcher (postdoc) at the University of Aizu since April 1, 2024. His research interests include AI, ML, DL, Human Activity Recognition (HCR), Hand Gesture Recognition (HGR), Movement Disorder Detection, Parkinson's Disease (PD), HCI, BCI, and Neurological Disorder Detection. He has authored and co-authored more than 50 publications in widely cited journals and conferences.
\end{IEEEbiography}

\begin{IEEEbiography}    [{\includegraphics[width=1in,height=1.25in,clip,keepaspectratio]{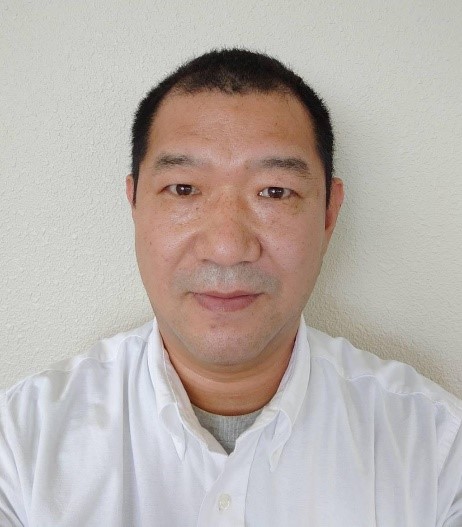}}]
{Nobuyoshi Asai} received his B.Eng. in Information Science, M. Eng. and Ph.D. in Engineering from the University of Tsukuba, Japan, in 1991, 1993, and 1996. He worked for Wave Front Co. Ltd., Japan, in 1996 and he was an Associate Professor and a Senior Associate Professor at the school of Computer Science and Engineering, the University of Aizu, Japan, from 2000 and from 2010. He was also a visiting researcher for National Institute of Environmental Study, Japan, in 2002 and 2003. His research interests include Applied Math., Numerical Analysis, Quantum Algorithms, High Performance Computing, Synesthesia, Signal Processing on EEG or Motion, etc. He is a member of JSIAM and IPSJ. He served an editor for the Journal of IPSJ, and currently, serves as Editor in Chief. He also served as a general chair for an IEEE international conference.
\end{IEEEbiography}
\begin{IEEEbiography}[{\includegraphics[width=1in,height=1.25in,clip,keepaspectratio]{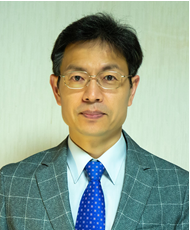}}]{JUNGPIL SHIN}(Senior Member, IEEE) received the B.Sc. degree in computer science and statistics and the M.Sc. degree in computer science from Pusan National University, South Korea, in 1990 and 1994, respectively, and the Ph.D. degree in computer science and communication engineering from Kyushu University, Japan, in 1999, under a scholarship from the Japanese Government (MEXT). He was an Associate Professor, a Senior Associate Professor, and a Full Professor with the School of Computer Science and Engineering, The University of Aizu, Japan, in 1999, 2004, and 2019, respectively. He has coauthored more than 420 published papers for widely cited journals and conferences. His research interests include pattern recognition, image processing, computer vision, machine learning, human–computer interaction, non-touch interfaces, human gesture recognition, automatic control, Parkinson’s disease diagnosis, ADHD diagnosis, user authentication, machine intelligence, bioinformatics, and handwriting analysis, recognition, and synthesis. He is a member of ACM, IEICE, IPSJ, KISS, and KIPS. He serves as an Editorial Board Member for Scientific Reports. He was included among the top 2\% of scientists worldwide edition of Stanford University/Elsevier, in 2024. He served as the general chair, the program chair, and a committee member for numerous international conferences. He serves as an Editor for IEEE journals, Springer, Sage, Taylor \& Francis, Sensors (MDPI), Electronics (MDPI), and Tech Science. He serves as a reviewer for several major IEEE and SCI journals.

\end{IEEEbiography}

\EOD

\end{document}